\documentclass{article}

\usepackage{microtype}
\usepackage{graphicx}
\usepackage{subfigure}
\usepackage{booktabs} 

\usepackage{hyperref}

\usepackage[accepted]{icml2019}

\usepackage[utf8]{inputenc}
\usepackage{comment,url,algorithm,algorithmic,graphicx,relsize} 
\usepackage{amssymb,amsfonts,amsmath,amsthm,amscd,dsfont,mathrsfs,mathtools,nicefrac}
\usepackage{float,psfrag,epsfig,color,xcolor,url,hyperref}
\usepackage{epstopdf,bbm,mathtools,enumitem}

\def\balign#1\ealign{\begin{align}#1\end{align}}
\def\baligns#1\ealigns{\begin{align*}#1\end{align*}}
\def\balignat#1\ealign{\begin{alignat}#1\end{alignat}}
\def\balignats#1\ealigns{\begin{alignat*}#1\end{alignat*}}
\def\bitemize#1\eitemize{\begin{itemize}#1\end{itemize}}
\def\benumerate#1\eenumerate{\begin{enumerate}#1\end{enumerate}}

\newenvironment{talign*}
 {\csname align*\endcsname}
 {\endalign}
\newenvironment{talign}
 {\csname align\endcsname}
 {\endalign}

\def\balignst#1\ealignst{\begin{talign*}#1\end{talign*}}
\def\balignt#1\ealignt{\begin{talign}#1\end{talign}}

\let\originalleft\left
\let\originalright\right
\renewcommand{\left}{\mathopen{}\mathclose\bgroup\originalleft}
\renewcommand{\right}{\aftergroup\egroup\originalright}

\def\tinycitep*#1{{\tiny\citep*{#1}}}
\def\tinycitealt*#1{{\tiny\citealt*{#1}}}
\def\tinycite*#1{{\tiny\cite*{#1}}}
\def\smallcitep*#1{{\scriptsize\citep*{#1}}}
\def\smallcitealt*#1{{\scriptsize\citealt*{#1}}}
\def\smallcite*#1{{\scriptsize\cite*{#1}}}

\def\mbb#1{\mathbb{#1}}

\def\<{\left\langle} 
\def\>{\right\rangle}

\def\E{\mbb{E}} 

\providecommand{\argmin}{\mathop\mathrm{arg min}}

\ifdefined\nonewproofenvironments\else
\ifdefined\ispres\else

\newenvironment{proof-sketch}{\noindent\textbf{Proof Sketch}
  \hspace*{1em}}{\qed\bigskip\\}
\newenvironment{proof-idea}{\noindent\textbf{Proof Idea}
  \hspace*{1em}}{\qed\bigskip\\}
\newenvironment{proof-of-lemma}[1][{}]{\noindent\textbf{Proof of Lemma {#1}}
  \hspace*{1em}}{\qed\\}
\newenvironment{proof-of-theorem}[1][{}]{\noindent\textbf{Proof of Theorem {#1}}
  \hspace*{1em}}{\qed\\}
\newenvironment{proof-attempt}{\noindent\textbf{Proof Attempt}
  \hspace*{1em}}{\qed\bigskip\\}

\fi

\fi
\makeatletter
\@addtoreset{equation}{section}
\makeatother

\hypersetup{
  colorlinks,
  linkcolor={red!50!black},
  citecolor={blue!50!black},
  urlcolor={blue!80!black}
}

\newcommand{\handout}[5]{
  \noindent
  \begin{center}
    \framebox{
      \vbox{
        \hbox to 5.78in { {\bf \title } \hfill #2 }
        \vspace{4mm}
        \hbox to 5.78in { {\Large \hfill #5  \hfill} }
        \vspace{2mm}
        \hbox to 5.78in { {\em #3 \hfill #4} }
      }
    }
  \end{center}
  \vspace*{4mm}
}

\mathtoolsset{showonlyrefs=true}

\usepackage{tikz}
\usetikzlibrary{positioning}

\newcommand{\Real}{\mathbb{R}}  

\newcommand{\eparam}{\boldsymbol{\theta}}  
\newcommand{\dimsymbol}{\textnormal{d}} 
\newcommand{\edom}{\Theta}  
\newcommand{\vx}{\boldsymbol{\textnormal{x}}}  
\newcommand{\xdim}{\dimsymbol_{\vx}}  
\newcommand{\vy}{\boldsymbol{\textnormal{y}}}  
\newcommand{\ydim}{\dimsymbol_{\vy}}  
\newcommand{\vpred}{\hat{\vy}_{\eparam}}  
\newcommand{\xdom}{\mathcal{X}}  
\newcommand{\ydom}{\mathcal{Y}}  
\newcommand{\dataset}{\mathcal{D}}  
\newcommand{\datanum}{n}  
\newcommand{\loss}{\mathcal{L}}  
\newcommand{\risk}{R}  
\newcommand{\emprisk}{\hat{\risk}}  
\newcommand{\jacy}{J^{\vy}_{\vx}} 
\newcommand{\predjacy}{J_{\eparam}} 
\newcommand{\xpath}{\boldsymbol{c}} 
\newcommand{\dxpath}{\xpath'} 
\newcommand{\vxo}{\vx_{o}}  
\newcommand{\vyo}{\vy_{o}}  
\newcommand{\pvx}{p(\vx)}  

\icmltitlerunning{JacNet: Learning Functions with Structured Jacobians}

\begin{document}

\twocolumn[
    \icmltitle{JacNet: Learning Functions with Structured Jacobians}
    
    
    
    \icmlsetsymbol{equal}{*}
    
    \begin{icmlauthorlist}
    \icmlauthor{Jonathan Lorraine}{equal,to,vec}
    \icmlauthor{Safwan Hossain}{equal,to,vec}
    \end{icmlauthorlist}
    
    \icmlaffiliation{to}{University of Toronto}
    \icmlaffiliation{vec}{Vector Institute}
    
    \icmlcorrespondingauthor{Jonathan Lorraine}{lorraine@cs.toronto.edu}
    
    \icmlkeywords{Machine Learning, ICML}
    
    \vskip 0.15in
]



\printAffiliationsAndNotice{\icmlEqualContribution} 

\begin{abstract}
    Neural networks are trained to learn an approximate mapping from an input domain to a target domain.
    Incorporating prior knowledge about true mappings is critical to learning a useful approximation.
    With current architectures, it is challenging to enforce structure on the derivatives of the input-output mapping.
    We propose to use a neural network to directly learn the Jacobian of the input-output function, which allows easy control of the derivative.
    We focus on structuring the derivative to allow invertibility and also demonstrate that other useful priors, such as $k$-Lipschitz, can be enforced.
    Using this approach, we can learn approximations to simple functions that are guaranteed to be invertible and easily compute the inverse.
    We also show similar results for 1-Lipschitz functions.
\end{abstract}

    
    
    
    
    
    
    
    

\section{Introduction}
    Neural networks (NNs) are the main workhorses of modern machine learning, used to approximate functions in a wide range of domains.
    Two traits that drive NN's success are (1) they are sufficiently flexible to approximate arbitrary functions, and (2) we can easily structure the output to incorporate certain prior knowledge about the range (e.g., softmax output activation for classification).
    
    NN flexibility is formalized by showing they are \emph{universal function approximators}.
    This means that given a continuous function $\vy$ on a bounded interval $I$, a NN approximation $\vpred$ can satisfy: $\forall \vx \in I, |\vy(\vx) - \vpred(\vx)| < \epsilon$.
    \citet{hornik1989multilayer} show that NNs with one hidden layer and non-constant, bounded activation functions are universal approximators.
    While NNs can achieve point-wise approximations with arbitrary precision, less can be said about their derivatives w.r.t. the input.
    For example, NNs with step-function activations are flat almost everywhere and can not approximate arbitrary input derivatives yet are still universal approximators.
    
    
    The need to use derivatives of approximated functions arises in many scenarios \cite{vicol2022implicit}.
    For example, in generative adversarial networks~\citep{goodfellow2014generative}, the generator differentiates through the discriminator to ensure the learned distribution is closer to the true distribution.
    In some multi-agent learning algorithms, an agent differentiates through how another agent responds \citep{foerster2018learning, lorraine2021lyapunov, lorraine2022complex}.
    Alternatively, hyperparameters can differentiate through a NN to see how to move to minimize validation loss \citep{mackay2019self, lorraine2018stochastic, lorraine2020optimizing, lorraine2022task, lorraine2024scalable, mehta2024improving, raghu2021meta, adam2019understanding, bae2024training, zhang2023using}.
    
        We begin by giving background on the problem in \S~\ref{sec:background} and discuss related work in \S~\ref{sec:related-work}.
        Then, we introduce relevant theory for our algorithm in \S~\ref{sec:theory} followed by our experimental results in \S~\ref{sec:experiments}.
    
    \subsection{Contributions}
        \begin{itemize}
            \item We propose a method for learning functions by learning their Jacobian, and provide empirical results.
            \item We show how to make our learned function satisfy regularity conditions - invertible, or Lipschitz - by making the Jacobian satisfy regularity conditions
            \item We show how the learned Jacobian can satisfy regularity conditions via appropriate output activations.
        \end{itemize}


\section{Background}\label{sec:background}
    This section sets up the standard notation used in \S~\ref{sec:theory}.
    Our goal is to learn a $C^1$ function $\vy(\vx): \xdom \to \ydom$.
    Denote $\xdim, \ydim$ as the dimension of $\xdom, \ydom$ respectively.
    Here, we assume that $\vx \sim \pvx$ and $\vy$ is deterministic.
    We will learn the function through a NN - $\vpred(\vx)$ - parameterized by weights $\eparam \in \edom$.
    Also, assume we have a bounded loss function $\loss(\vy(\vx), \vpred(\vx))$, which attains its minimum when $\vy(\vx) = \vpred(\vx)$.
    Our population risk is $\risk(\eparam) = \E_{\pvx}[\loss(\vy(\vx), \vpred(\vx))]$, and we wish to find $\eparam^{*} = \argmin_{\eparam} \risk(\eparam)$.
    In practice, we have a finite number of samples from the input distribution $\dataset = \{ (\vx_i, \vy_i) | i = 1 \dots \datanum \}$, and we minimize the empirical risk: 
    \begin{equation}
        \smash{\hat{\eparam}^{*} = \argmin_{\eparam} \emprisk(\eparam) = \argmin_{\eparam} \nicefrac{1}{n}\sum_{\dataset} \loss(\vy_i, \vpred(\vx_i, \eparam))}
    \end{equation}
    %
    It is common to have prior knowledge about the structure of $\vy(\vx)$ which we want to bake into the learned function.
    If we know the bounds of the output domain, properly structuring the predictions through output activation is an easy way to enforce this.
    Examples include using softmax for classification, or ReLU for a non-negative output.
    
    In the more general case, we may want to ensure our learned function satisfies certain derivative conditions, as many function classes can be expressed in such a way.
    For example, a function is locally invertible if its Jacobian has a non-zero determinant in that neighborhood.
    Similarly, a function is $k$-Lipschitz if its derivative norm lies inside $[-k, k]$.
    
    We propose explicitly learning this Jacobian through a NN $\predjacy(\vx)$ parameterized by $\theta$ and using a numerical integrator to evaluate $\vpred(\vx)$.
    We show that with a suitable choice of output activation for $\predjacy(\vx)$, we can guarantee our function is globally invertible, or $k$-Lipschitz.



\section{Related Work}\label{sec:related-work}
    
    \textbf{Enforcing derivative conditions}:
        There is existing work on strictly enforcing derivative conditions on learned functions.
        For example, if we know the function to be $k$-Lipschitz, one method is weight clipping~\citep{arjovsky2017wasserstein}.
        \citet{anil2018sorting} recently proposed an architecture which learns functions that are guaranteed to be 1-Lipschitz and theoretically capable of learning all such functions.
        \citet{amos2017input} explore learning scalar functions that are guaranteed to be convex (i.e., the Hessian is positive semi-definite) in their inputs.
        While these methods guarantee derivative conditions, they can be non-trivial in generalizing to new conditions, limiting expressiveness, or involving expensive projections.
        \citet{czarnecki2017sobolev} propose a training regime that penalizes the function when it violates higher-order constraints.
        This does not guarantee the regularity conditions and requires knowing the exact derivative at each sample - however, it is easy to use.

    \textbf{Differentiating through integration}:
        Training our proposed model requires back-propagating through a numerical integration procedure.
        We leverage differentiable numerical integrators provided by \citet{neuralODE} who use it to model the dynamics of different layers of an NN as ordinary differential equations.
        FFOJRD~\citep{grathwohl2018ffjord} uses this for training as it models layers of a reversible network as an ODE.
        Our approach differs in that we explicitly learn the Jacobian of our input-output mapping and integrate along arbitrary paths in the input or output domain.
        
        
    \textbf{Invertible Networks}:
        In \citet{behrmann2018invertible}, an invertible residual network is learned with contractive layers, and a numerical fixed point procedure is used to evaluate the inverse.
        In contrast, we need a non-zero Jacobian determinant and use numerical integration to evaluate the inverse.
        Other reversible architectures include NICE~\citep{dinh2014nice}, Real-NVP~\citep{dinh2016density}, RevNet~\citep{jacobsen2018revnet}, and Glow~\citep{kingma2018glow}.
        \citet{richter2021input} develop continuations on our workshop method \cite{lorraine2019jacnet}.

\section{Theory}\label{sec:theory}
    We are motivated by the idea that a function can be learned by combining initial conditions with an approximate Jacobian.
    Consider a deterministic $C^1$ function $\vy: \xdom \to \ydom$ that we wish to learn.
    Let $\jacy:\xdom \to \Real^{\xdim \times \ydim}$ be the Jacobian of $\vy$ w.r.t. $\vx$.
    We can evaluate the target function by evaluating the following line integral with some initial condition $(\vxo, \vyo = \vy(\vxo))$:
    \begin{equation}
        \vy(\vx) = \vyo + \smash{\int_{\xpath(\vxo, \vx)} \jacy(\vx) ds}
    \end{equation}
    In practice, this integral is evaluated by parameterizing a path between $\vxo$ and $\vx$ and numerically approximating the integral.
    Note that the choice of path and initial condition do not affect the result by the fundamental theorem of line integrals.
    We can write this as an explicit line integral for some path $\xpath(t, \vxo, \vx)$ from $\vxo$ to $\vx$ parameterized by $t \in [0, 1]$ satisfying $\xpath(0, \vxo, \vx) = \vxo$ and $\xpath(1, \vxo, \vx) = \vx$ with $\nicefrac{d}{dt}(\xpath(t, \vxo, \vx)) = \dxpath(t, \vxo, \vx)$:
    \begin{equation}
        \vy(\vx) = \vyo + \smash{\int_{t=0}^{t=1}{\jacy(\xpath(t, \vxo, \vx))\dxpath(t, \vxo, \vx)dt}}
    \end{equation}
    A simple choice of path is $\xpath(t, \vxo, \vx) = (1-t)\vxo + t\vx$, which has $\dxpath(t, \vxo, \vx) = \vx - \vxo$.
    Thus, to approximate $\vy(\vx)$, we can combine initial conditions with an approximate $\jacy$.
    We propose to learn an approximate, Jacobian $\predjacy(\vx): \xdom \to \Real^{\xdim \times \ydim}$, with a NN parameterized by $\eparam \in \edom$.
    For training, consider the following prediction function:
    \begin{equation}
        \vpred(\vx) = \vyo + \smash{\int_{t=0}^{t=1}{\predjacy(\xpath(t, \vxo, \vx))\dxpath(t, \vxo, \vx)dt}}
    \end{equation}
    We can compute the empirical risk, $\emprisk$, with this prediction function by choosing some initial conditions $(\vxo, \vyo) \in \dataset$, a path $\xpath$, and using numerical integration.
    To backpropagate errors to update our network parameters $\eparam$, we must backpropagate through the numerical integration.
    
    \subsection{Derivative Conditions for Invertibility}
        The inverse function theorem~\citep{spivak2018calculus} states that a function is locally invertible if the Jacobian at that point is invertible.
        Additionally, we can compute the Jacobian of $f^{-1}$ by computing the inverse of the Jacobian of $f$.
        Many non-invertible functions are locally invertible almost everywhere (e.g., $y = x^2$).
        
        The Hadamard global inverse function theorem~\citep{hadamard1906transformations}, is an example of a global invertibility criterion.
        It states that a function $f: \xdom \to \ydom$ is globally invertible if the Jacobian determinant is everywhere non-zero and $f$ is \textit{proper}.
        A function is \textit{proper} if whenever $\ydom$ is compact, $f^{-1}(\ydom)$ is compact.
        This provides an approach to guarantee global invertibility of a learned function.
    
    \subsection{Structuring the Jacobian}
        By guaranteeing it has non-zero eigenvalues, we could guarantee that our Jacobian for an $\Real^n \to \Real^n$ mapping is invertible.
        For example, with a small positive $\epsilon$ we could use an output activation of:
        \begin{equation}
            \smash{\predjacy'(\vx) = \predjacy(\vx) \predjacy^{T}(\vx) + \epsilon I}
        \end{equation}
        Here, $\predjacy(\vx) \predjacy^{T}(\vx)$ is a flexible PSD matrix, while adding $\epsilon I$ makes it positive definite.
        A positive definite matrix has strictly positive eigenvalues, which implies invertibility.
        However, this output activation restricts the set of invertible functions we can learn, as the Jacobian can only have positive eigenvalues.
        In future work, we wish to explore less restrictive activations while still guaranteeing invertibility.
        
        Many other regularity conditions on a function can be specified in terms of the derivatives.
        For example, a function is Lipschitz if the function's derivatives are bounded, which can be done with a $k$-scaled $\tanh$ activation function as our output activation.
        Alternatively we could learn a complex differentiable function by satisfying $\nicefrac{\partial u}{\partial a} = \nicefrac{\partial v}{\partial b}, \nicefrac{\partial u}{\partial b} = -\nicefrac{\partial v}{\partial a}$, where $u ,v$ are output components and $a, b$ are input components.
        We focus on invertibility and Lipschitz because they are common in machine learning.
    
    \subsection{Computing the Inverse}
        Once the Jacobian is learned, it allows easy computation of $f^{-1}$ by integrating the inverse Jacobian along a path $\xpath(t, \vyo, \vy)$ in the output space, given some initial conditions $(\vxo, \vyo = \vpred(\vxo))$:
        \begin{equation}
            \vx(\vy, \eparam) = \vxo + \smash{\int_{t=0}^{t=1}{\left(\predjacy(\xpath(t, \vyo, \vy))\right)^{-1}\dxpath(t, \vyo, \vy)dt}}
        \end{equation}
        If the computational bottleneck is inverting $\predjacy$, we propose to learn a matrix which is easily invertible (e.g., Kronecker factors of $\predjacy$).

         
    \subsection{Backpropagation}
        Training the model requires back-propagating through numerical integration.
        To address this, we consider the recent work of \citet{neuralODE}, which provides tools to efficiently back-propagate across numerical integrators.
        We combine their differentiable integrators on intervals with autograd for our rectification term $\dxpath$.
        This provides a differentiable numerical line integrator, which only requires a user to specify a differentiable path and the Jacobian.
        
        The integrators allow a user to specify solution tolerance.
        We propose annealing the tolerance tighter whenever the evaluated loss is lower than our tolerance.
        In practice, this provides significant computational savings.
        Additionally, suppose the computational bottleneck is numerical integration. In that case, we may be able to adaptively select initial conditions $(\vxo, \vyo)$ that are near our target, reducing the number of evaluation steps in our integrator.
        
    \subsection{Conservation}
        When our input domain dimensionality $\xdim > 1$, we run into complexities.
        For simplicity, assume $\ydim = 1$, and we are attempting to learn a vector field that is the gradient.
        Vector fields that are the gradient of a function are known as conservative.
        Our learned function $\predjacy$ is a vector field but is not necessarily conservative.
        As such, $\predjacy$ may not be the gradient of any scalar potential function, and the value of our line integral depends on the path choice.
        Investigating potential problems and solutions to these problems is relegated to future work.

\section{Experiments}\label{sec:experiments}
    \newcommand{\expNumTrain}{5}  
    \newcommand{\batchSize}{1}
    \newcommand{\trainDom}{[-1, 1]}
    \newcommand{\expNumPopEval}{100}
    \newcommand{\invFunc}{\exp}
    \newcommand{\testDom}{[-2, 2]}
    \newcommand{\numLayers}{1}
    \newcommand{\numHidden}{64}
    \newcommand{\learningRate}{0.01}
    \newcommand{\dampFactor}{0.0001}
    \newcommand{\numEpoch}{50}
    In our experiments, we explore learning invertible, and Lipschitz functions with the following setup:
    Our input and output domains are $\xdom = \ydom = \Real$.
    We select $\loss(\vy_1, \vy_2) = \|\vy_1 - \vy_2 \|$.
    Our training set consists of $\expNumTrain$ points sampled uniformly from $\trainDom$, while our test set has $\expNumPopEval$ points sampled uniformly from $\testDom$.
    The NN architecture is fully connected with a single layer with $\numHidden$ hidden units, and output activation on our network depends on the gradient regularity condition we want.
    We used Adam~\citep{kingma2014adam} to optimize our network with a learning rate of $\learningRate$ and all other parameters at defaults.
    We use full batch gradient estimates for $\numEpoch$ iterations.
    
    To evaluate the function at a point, we use an initial value, parameterize a linear path between the initial and terminal point, and use the numerical integrator from \citet{neuralODE}.
    The path is trivially the interval between $\vxo$ and $\vx$ because $\xdim = 1$, and our choice of the initial condition is $\vxo = 0$.
    We adaptively alter the integrator's tolerances, starting with loose tolerances and decreasing them by half when the training loss is less than the tolerance, which provides significant gains in training speed.
    
\begin{figure}
        \vspace{-.25cm}
        \centering
        \begin{tikzpicture}
            \centering
            \node (img){\includegraphics[width=0.8\linewidth]{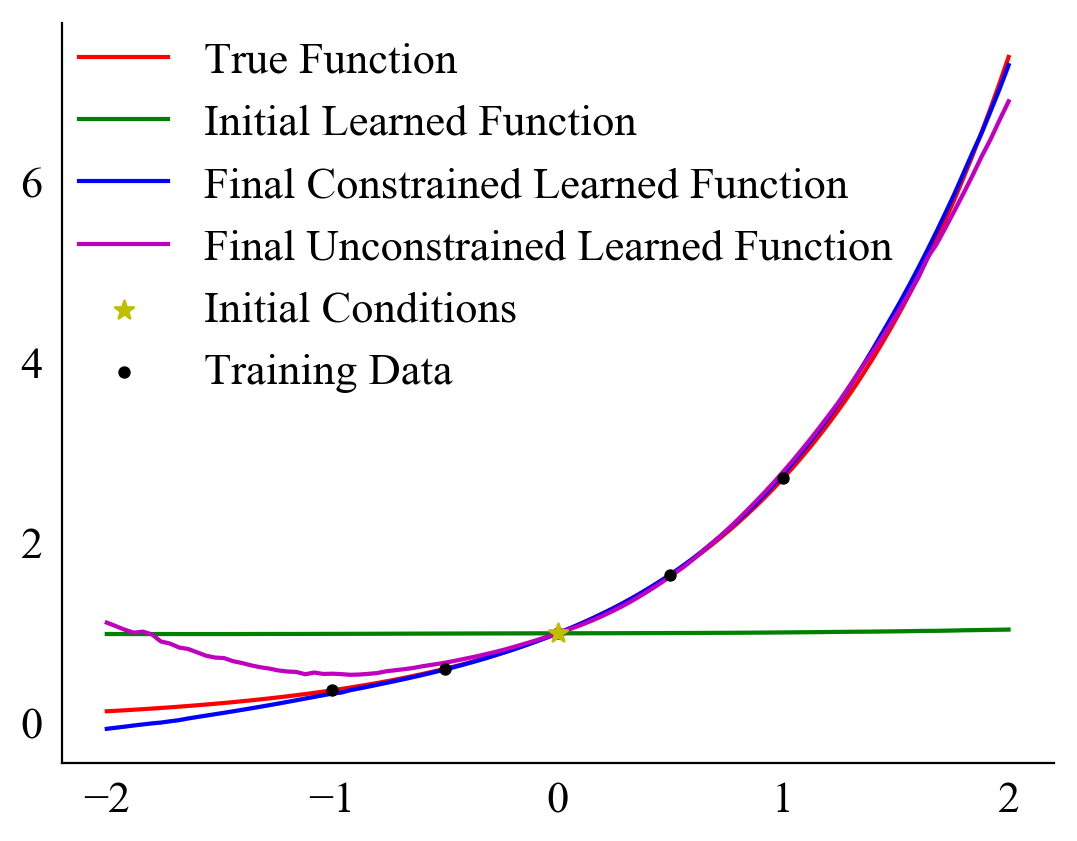}};
            \node[left=of img, node distance=0cm, rotate=90, xshift=1.75cm, yshift=-.75cm, font=\color{black}] {Target $\vy$ / Prediction $\vpred$};
            \node[below=of img, node distance=0cm, yshift=1.25cm,font=\color{black}] {Input $\vx$};
            
            \node (img2)[below=of img, node distance=0cm, yshift=.75cm, xshift=-.25cm]{\includegraphics[width=0.8\linewidth]{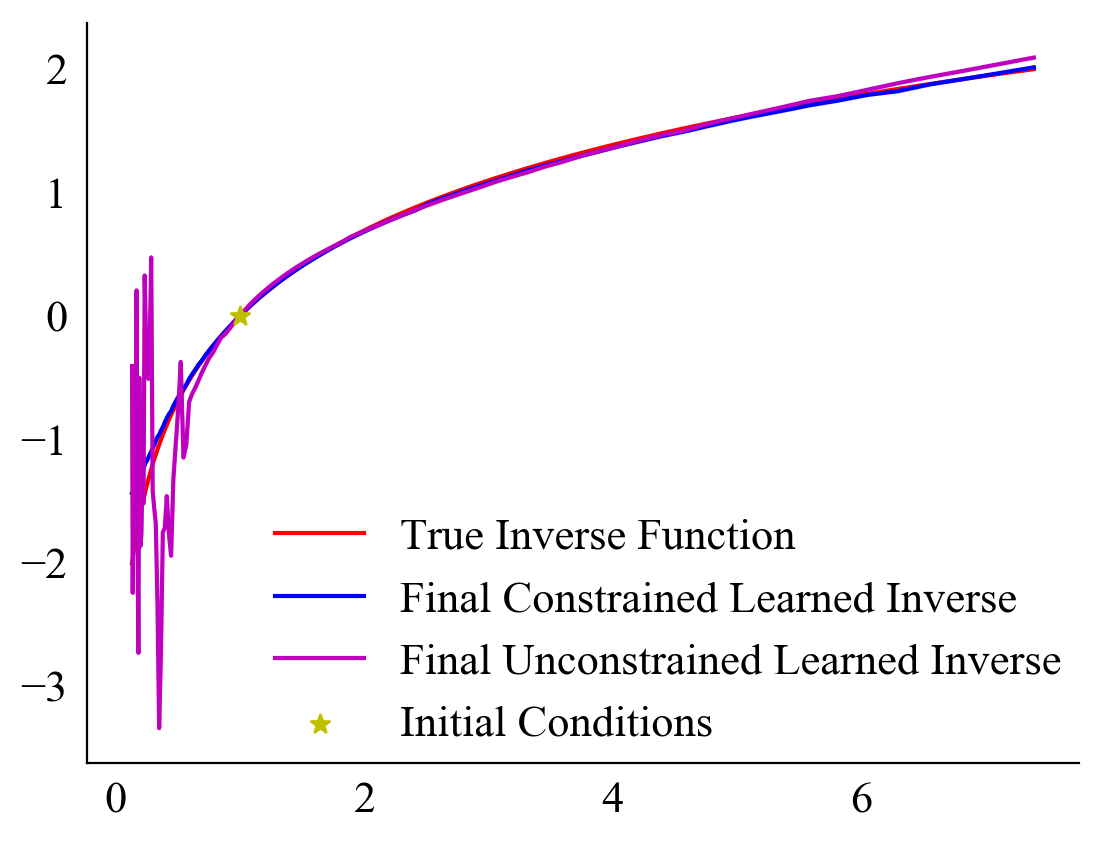}};
            \node[left=of img2, node distance=0cm, rotate=90, xshift=1.25cm, yshift=-1cm, font=\color{black}] {Inverted $\vx$};
            \node[below=of img2, node distance=0cm, yshift=1.25cm, xshift=.25cm, font=\color{black}] {Input $\vy$};
        \end{tikzpicture}
        \vspace{-.5cm}
        \caption{
            A graph of the \emph{invertible} target function $\invFunc(\vx)$, the prediction function at the trained network weights, and the prediction function at the initial network weights.
            The initial prediction function is inaccurate, while the final one matches the target function closely.
            Furthermore, the learned inverse of the prediction function closely matches the true inverse function.
            We include graphs of an unconstrained learned function whose Jacobian can be zero.
            Note that the unconstrained function is not invertible everywhere.
        }
        \label{fig:experiment_invertible_compare}
        \vspace{-1cm}
    \end{figure}
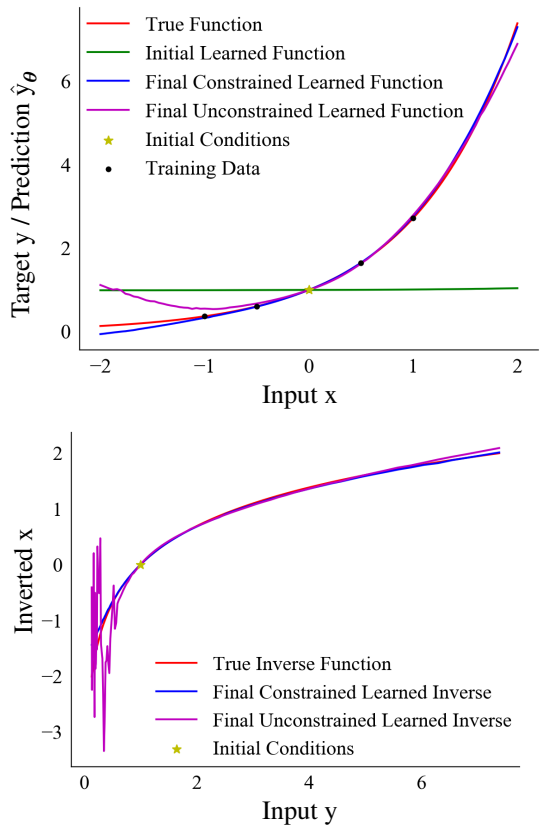

\begin{figure}
        \vspace{-.3cm}
        \centering
        \begin{tikzpicture}
            \centering
            \node (img) {\includegraphics[width=0.8\linewidth]{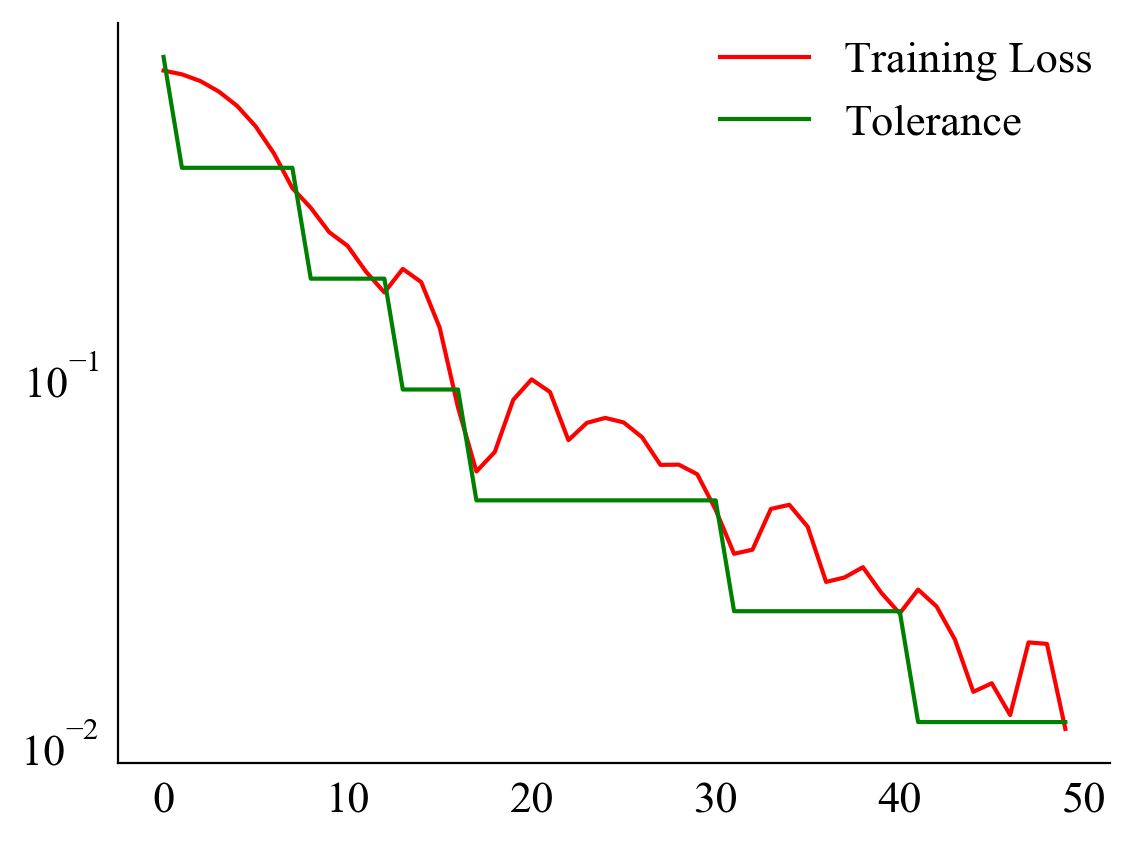}};
            \node[left=of img, node distance=0cm, rotate=90, xshift=1.5cm, yshift=-.8cm,font=\color{black}] {Empirical Risk $\emprisk(\eparam)$};
            \node[above=of img, node distance=0cm, yshift=-1cm,font=\color{black}] {\emph{Top}: Learning invertible $\invFunc(\vx)$};
            \node[below=of img, node distance=0cm, yshift=1cm,font=\color{black}] {\emph{Bottom}: Learning Lipschitz $| \vx |$};
            
            \node [below=of img, node distance=0cm, yshift=.5cm] (img2) {\includegraphics[width=0.8\linewidth]{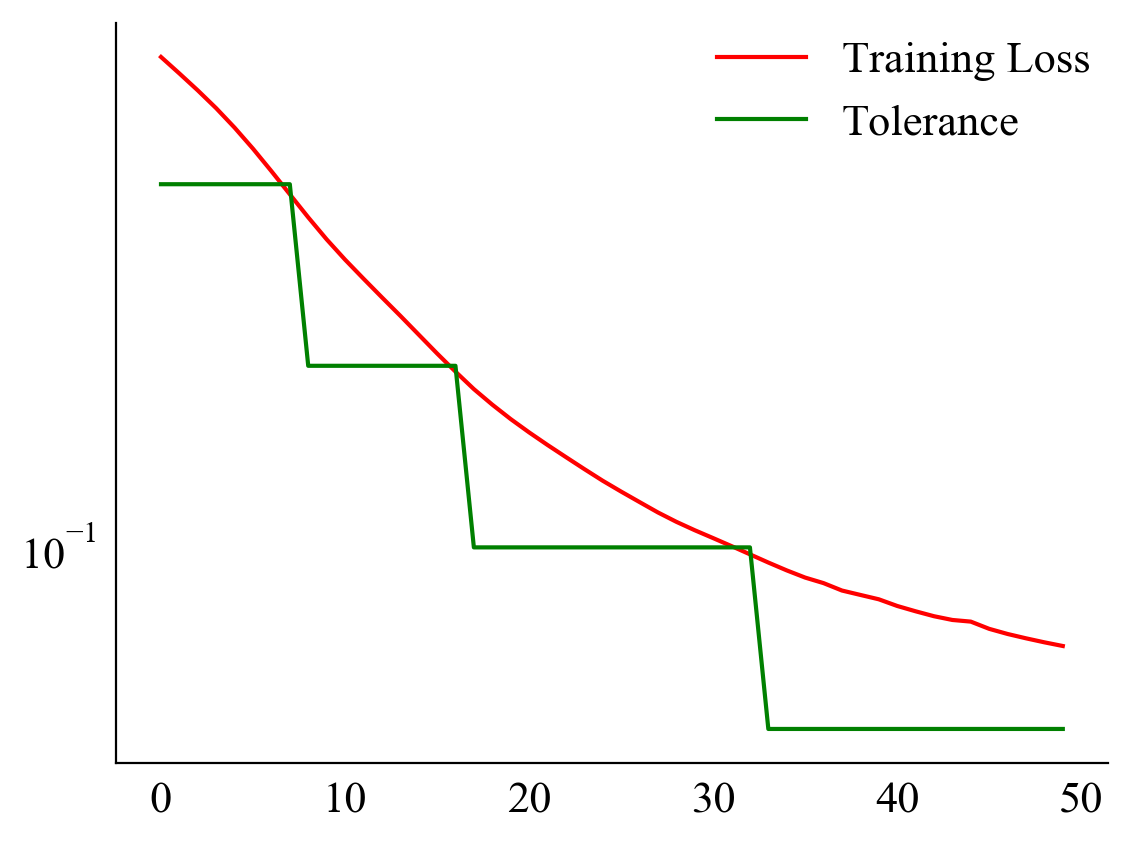}};
            \node[left=of img2, node distance=0cm, rotate=90, xshift=1.5cm, yshift=-.8cm,font=\color{black}] {Empirical Risk $\emprisk(\eparam)$};
            \node[below=of img2, node distance=0cm, yshift=1.25cm,font=\color{black}] {Iteration};
        \end{tikzpicture}
        \vspace{-.4cm}
        \caption{
            A graph of the empirical risk or training loss versus training iteration.
            As training progresses, we tighten the tolerance of the numerical integrator to continue decreasing the loss at the cost of more computationally expensive iterations.
            \emph{Top}: The training dynamics for learning the invertible target function.
            \emph{Bottom}: The training dynamics for learning the Lipschitz target function.
        }
        
        \label{fig:experiment_loss}
    \end{figure}

    We learn the exponential function $\vy(\vx) = \invFunc(\vx)$ for the invertible function experiment.
    We use an output activation of $\predjacy'(\vx) = \predjacy(\vx) \predjacy(\vx)^{T} + \epsilon I$ for $\epsilon = \dampFactor$, which guarantees a non-zero Jacobian determinant.
    Once trained, we take the learned derivative $\predjacy'(\vx)$ and compute its inverse, which by the inverse function theorem gives the derivative of the inverse.
    We compute the inverse of the prediction function by integrating the inverse of the learned derivative.
    Figure~\ref{fig:experiment_invertible_compare} qualitatively explores the learned function, and the top of Figure~\ref{fig:experiment_loss} quantitatively explores the training procedure.
    
\begin{figure}
        \vspace{-.25cm}
        \centering
        \begin{tikzpicture}
            \centering
            \node (img) {\includegraphics[width=0.8\linewidth]{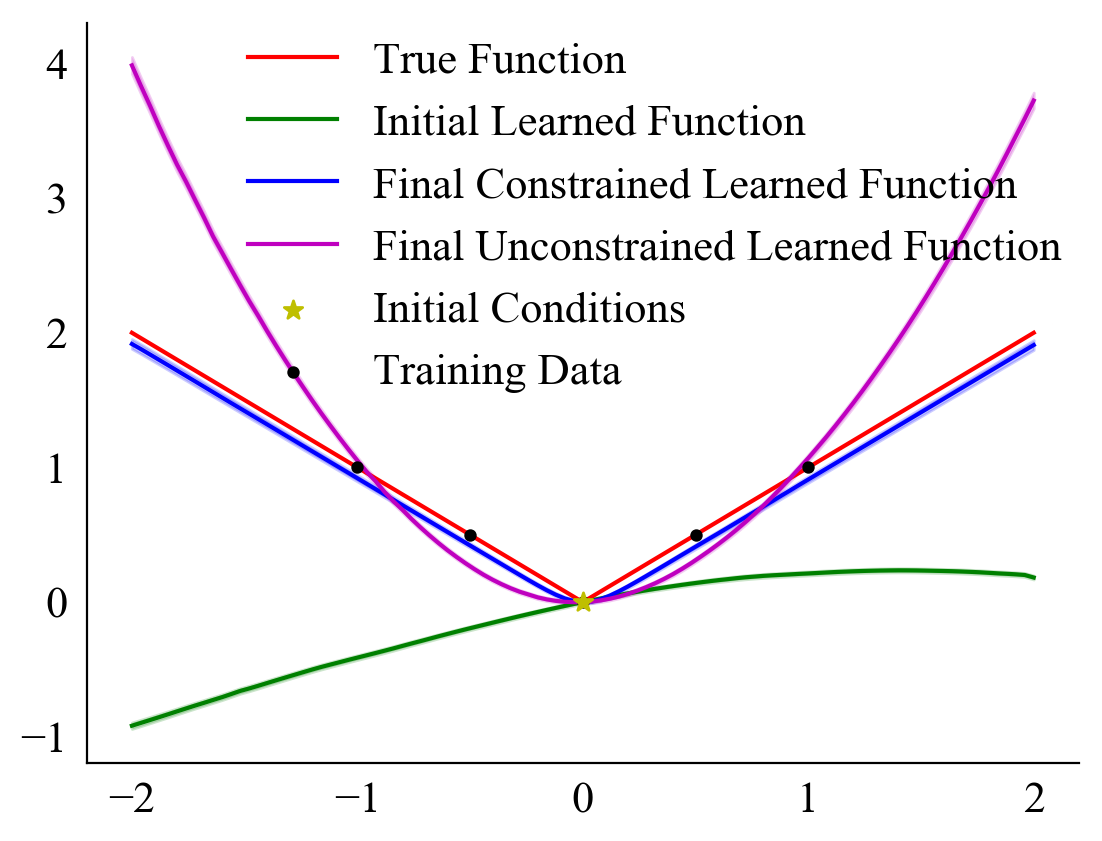}};
            \node[left=of img, node distance=0cm, rotate=90, xshift=1.75cm, yshift=-1cm, font=\color{black}] {Target $\vy$ / Prediction $\vpred$};
            \node[below=of img, node distance=0cm, yshift=1.25cm,font=\color{black}] {Input $\vx$};
        \end{tikzpicture}
        \vspace{-.5cm}
        \caption{
            A graph of the \emph{1-Lipschitz} target function $|\vx|$, the prediction function at the trained network weights, and the prediction function at the initial network weights.
            Note how the initial prediction function is inaccurate while the final prediction function matches the target function closely.
            Additionally, note how the learned prediction function is 1-Lipschitz at initialization and after training.
            We include graphs of an unconstrained learned function whose derivative is not bounded by [-1, 1]. This does not generalize well to unseen data.
        }
        \label{fig:experiment_lipschitz_compare}
    \end{figure}

    For the Lipschitz experiment, we learn the absolute value function $\vy(\vx) = |\vx|$, a canonical 1-Lipschitz example in \citet{anil2018sorting}.
    We use an output activation on our NN of $\predjacy'(\vx) = \tanh(\predjacy(\vx)) \in [-1, 1]$, which guarantees our prediction function is 1-Lipschitz.
    Figure~\ref{fig:experiment_lipschitz_compare} qualitatively explores the learned function, and the bottom of Figure~\ref{fig:experiment_loss} quantitatively explores the training procedure.
    
\section{Conclusion}\label{sec:conclusion}
    We present a technique to approximate a function by learning its Jacobian and integrating it.
    This method is useful when guaranteeing the function's Jacobian properties.
    A few examples of this include learning invertible, Lipschitz, or complex differentiable functions.
    Small-scale experiments are presented, motivating further exploration to scale up the results.
    We hope that this work will facilitate domain experts' easy incorporation of a wide variety of Jacobian regularity conditions into their models in the future.


\bibliography{bibliography}
\bibliographystyle{icml2019}


\newpage

\end{document}